\documentclass[10pt,twocolumn,letterpaper]{article}

\usepackage[utf8]{inputenc} 
\usepackage[T1]{fontenc}    
\usepackage{booktabs}       
\usepackage{amsfonts}       
\usepackage{nicefrac}       
\usepackage{microtype}      
\usepackage{caption}
\usepackage{graphicx}
\usepackage{amsmath}
\usepackage{amssymb}
\usepackage{booktabs}
\usepackage{threeparttablex}
\usepackage{multirow}
\usepackage{algorithm}
\usepackage{algpseudocode}
\usepackage{diagbox}
\usepackage{adjustbox}
\usepackage{multirow}
\usepackage{breqn}
\usepackage{soul}
\usepackage{stmaryrd}
\usepackage{gradient-text}
\usepackage{fontawesome}
\usepackage{tcolorbox}
\usepackage{listings}
\usepackage{colortbl}
\usepackage[accsupp]{axessibility}

\usepackage{amssymb}
\usepackage{pifont}
\newcommand{\cmark}{\ding{51}}%
\newcommand{\xmark}{\ding{55}}%

\usepackage[pagenumbers]{cvpr} 
\definecolor{cvprblue}{rgb}{0.21,0.49,0.74}
\usepackage[pagebackref,breaklinks,colorlinks,allcolors=cvprblue]{hyperref}

\usepackage{setspace}
\def\paperID{} 
\def\confName{}
\def\confYear{}

\newcommand{\modelname}{Safe-Construct}

\title{Safe-Construct: Redefining Construction Safety Violation Recognition\\ as 3D Multi-View Engagement Task\vspace{-0.5em}}

\newcommand*{\affaddr}[1]{#1}
\author{%
Aviral Chharia$^{1}$,
Tianyu Ren$^{2}$,
Tomotake Furuhata$^{1}$,
Kenji Shimada$^{1}$\\
\affaddr{$^1$Carnegie Mellon University} \ \affaddr{$^2$University of Illinois Urbana-Champaign}\\
{\tt\small \{achharia, tomotake, shimada\}@andrew.cmu.edu, tianyur2@illinois.edu}\\
}
\begin{document}
\twocolumn[{
    \renewcommand\twocolumn[1][]{#1}
    \maketitle
    \begin{center}
    \centering
    \vspace{-1.5em}
    \includegraphics[width=\linewidth]{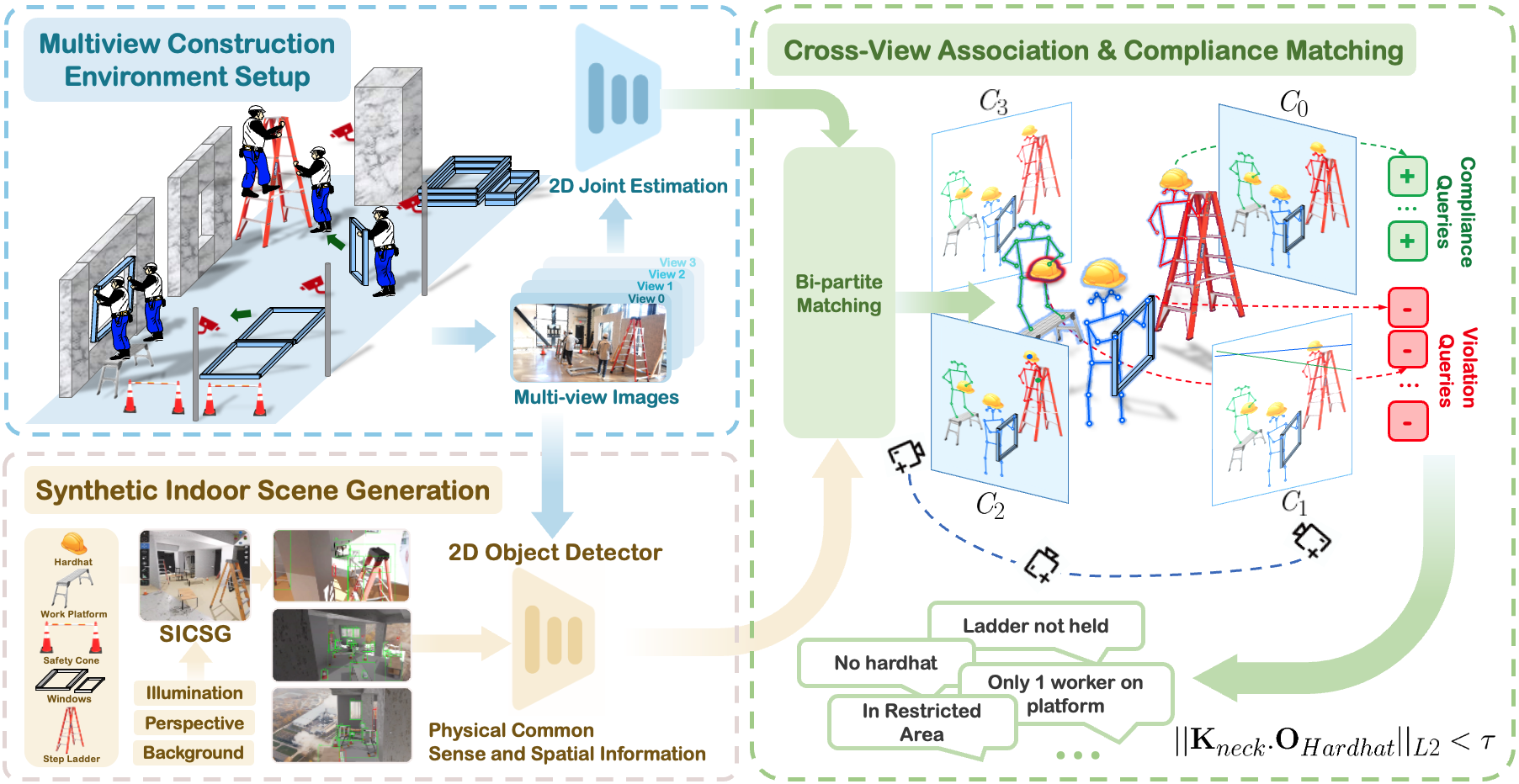}
    \captionof{figure}{\textbf{Overview of~\modelname{}.} We present~\modelname{}, the first 3D multi-view safety violation recognition model for construction sites. It consists of (a) a multi-camera setup, (b) a Synthetic Indoor Construction Site Generator (SICSG), and (c) a 3D cross-view association and compliance matching module. Please check the project page for video results.}
    \label{fig:teaser}
\end{center}}]
\begin{abstract}
\vspace{-2mm}
Recognizing safety violations in construction environments is critical yet remains underexplored in computer vision. Existing models predominantly rely on 2D object detection, which fails to capture the complexities of real-world violations due to: (i) an oversimplified task formulation treating violation recognition merely as object detection, (ii) inadequate validation under realistic conditions, (iii) absence of standardized baselines, and (iv) limited scalability from the unavailability of synthetic dataset generators for diverse construction scenarios. To address these challenges, we introduce~\modelname, the first framework that reformulates violation recognition as a 3D multi-view engagement task, leveraging scene-level worker-object context and 3D spatial understanding. We also propose the Synthetic Indoor Construction Site Generator (SICSG) to create diverse, scalable training data, overcoming data limitations. \modelname{} achieves a $7.6\%$ improvement over state-of-the-art methods across four violation types. We rigorously evaluate our approach in near-realistic settings, incorporating four violations, four workers, 14 objects, and challenging conditions like occlusions (worker-object, worker-worker) and variable illumination (back-lighting, overexposure, sunlight). By integrating 3D multi-view spatial understanding and synthetic data generation, \modelname{} sets a new benchmark for scalable and robust safety monitoring in high-risk industries. Project Website: \url{https://Safe-Construct.github.io/Safe-Construct}\\
\vspace{-8mm}
\end{abstract}
\makeatletter
\newcommand\figcaption{\def\@captype{figure}\caption}
\newcommand\tabcaption{\def\@captype{table}\caption}
\makeatother

\section{Introduction}
\label{sec:introduction}

Construction is one of the largest sectors of the economy, generating around USD 2.2 trillion~\cite{usconstruction2024} and employing over 8.2 million workers in $2024$--about 5\% of the total U.S. workforce)~\cite{uslabor2024}. Despite its scale, construction remains one of the most hazardous sectors, with $1056$ fatal accidents reported in the U.S. in 2022~\cite{uslabordeaths2022}. Nearly half of these deaths result from preventable causes, such as falls or being struck by objects~\cite{uslaborinjuries2019}. These statistics highlight the importance of following safety protocols, such as wearing protective gear and adhering to guidelines for equipment like work platforms and ladders. As a result, detecting safety violations has become a critical challenge in the industry.

Vision-based methods have shown promise in improving construction safety by detecting violations. However, existing models predominantly frame this problem as an object detection task, relying on oversimplified assumptions that limit their real-world applicability~\cite{park2015hardhat,delhi2020detection,nath2020deepfrontiers, nath2020deep, wu2019automatic,construction-site-safety_dataset,hayat2022deep,islam2024deep,jia2025geoiou,zhang2025detection,eum2025heavy,ding2025yolo,seth2025enhanced,shetty2024enhancing,chen2021detection,jiao2025detection,zan2025mkd,liu2025improved,onal2024unsafe,park2023small,praveena2025automated,sudharshan2025automated,hwang2025fd,chang2023ffa,jin2024yolo,lian2024hr,yan2022deep,yang2024yolo,cheng2024highly}. Recent works have attempted to move beyond this paradigm by incorporating skeletal data~\cite{xiong2021pose, bo2021skeleton} and modeling human-object interactions~\cite{tang2020human}. Yet, these methods rely solely on 2D inputs, which often fail under heavy occlusions common at construction sites. We identify several limitations of prior works:\\
\vspace{-0.75em}

\begin{figure}[t]
    \centering
    \includegraphics[width=\linewidth]{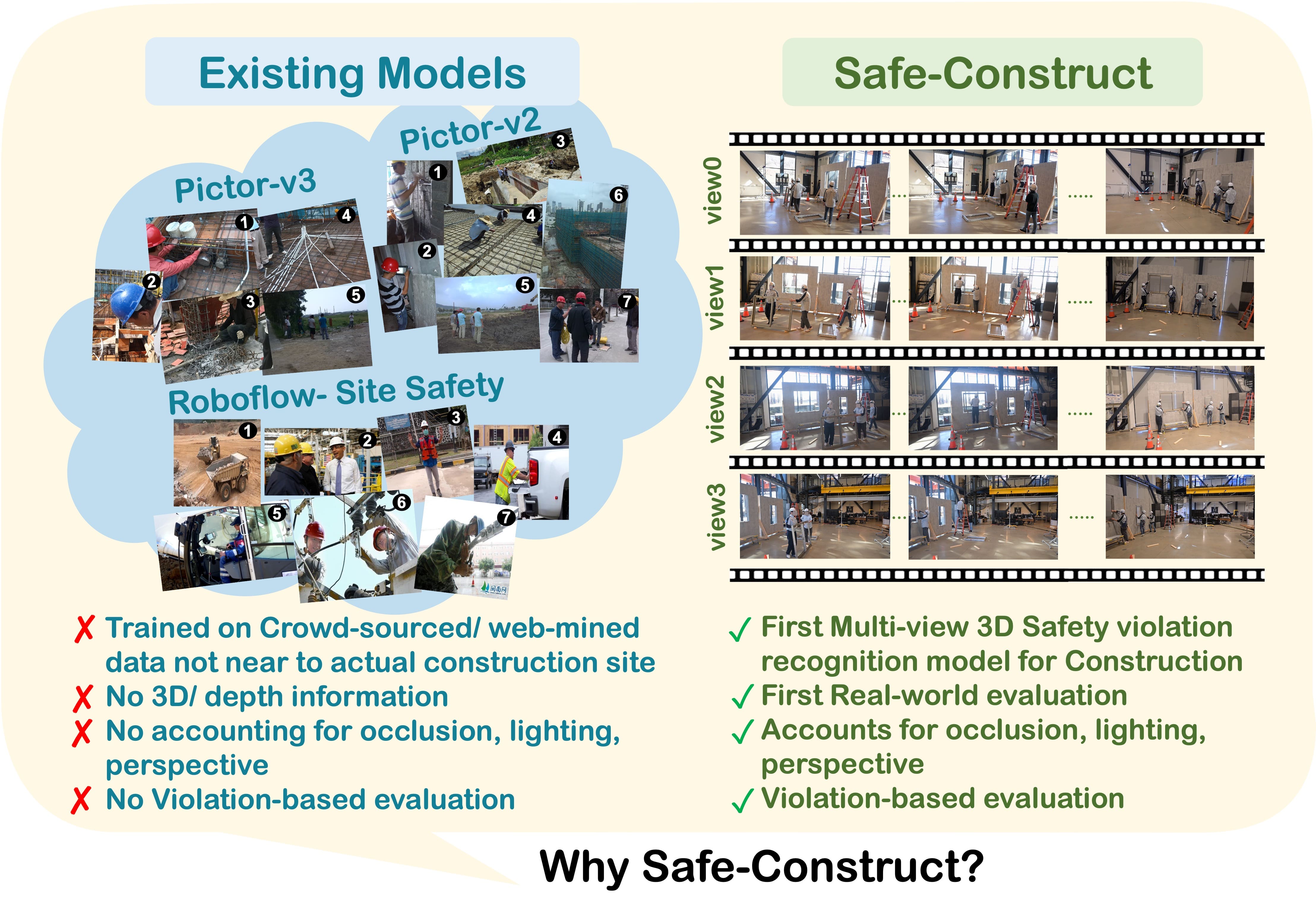}
    \captionof{figure}{\textbf{Comparison with Prior Methods.} Previous models~\cite{park2015hardhat,delhi2020detection,nath2020deepfrontiers, nath2020deep, wu2019automatic,construction-site-safety_dataset} train on crowd-sourced or web-mined data: Pictor-v2~\cite{nath2020deepfrontiers}, Pictor-v3~\cite{nath2020deep}, and Roboflow~\cite{construction-site-safety_dataset} framing the problem as 2D object detection task. These lack 3D spatial understanding and scene-level worker-object context. These datasets are small, with most images having unrealistic resolutions and perspectives (see pictures 1-4 in Pictor-v3, 1-2 in Pictor-v2, and 2-7 in Roboflow). Realistic industrial setups do not feature workers in close camera proximity ($<$$1$m). (b) In contrast, \modelname{} is the first multi-view 3D violation recognition model that leverages 3D spatial understanding and scene-level worker-object context.}
    \label{fig1:datahighlight}
\end{figure}

\noindent \textbf{Lack of Realistic Training Data.} Existing models~\cite{park2015hardhat,delhi2020detection,nath2020deepfrontiers, nath2020deep, wu2019automatic,construction-site-safety_dataset} are primarily trained on crowd-sourced or web-mined datasets, such as Pictor-v2~\cite{nath2020deepfrontiers}, Pictor-v3~\cite{nath2020deep}, and Roboflow~\cite{construction-site-safety_dataset}. This results in distributions that poorly reflect real-world construction environments (see Fig.~\ref{fig1:datahighlight}). They overlook perspective variations and camera distances, often assuming the camera is placed within 1 meter of the worker---an impractical setup on actual sites. Further, these datasets lack diversity in lighting conditions, such as back-lighting, indoor illumination, or over-exposure, making models less robust in real-world deployment. Further, they lack appropriate 3D spatial and depth information, which is critical for recognizing violations under occlusion---such as determining whether a ladder is stabilized by a second worker when the first climbs it or identifying if large windows are being handled by multiple workers. As a result, they often focus on overly simplistic scenarios like detecting hard hats or safety vests.\\
\vspace{-0.75em}

\noindent \textbf{Limited Scalability.} Existing models rely on predefined dataset categories (\textit{e.g.}, the presence of a ``hard hat'') to recognize violations. As a result, datasets must include category-level positive and negative image pairs to teach the model to identify violations as object categories. This tight coupling between violations and dataset classes limits scalability: detecting new types of violations requires collecting entirely new datasets, which is expensive, time-consuming, resource-intensive, and potentially an unsafe process.\\
\vspace{-0.75em}

\noindent \textbf{Lack of Scene-level Understanding and Worker-Object Context.} Existing approaches simplify safety violation recognition into object-centric tasks (\textit{e.g.}, detecting a ``hard hat''), neglecting broader scene-level understanding. Moreover, their training datasets typically contain isolated object images rather than full scenes, limiting the model's ability to capture contextual relationships between workers and objects, crucial for identifying intricate, real-world violations. As an example, Pictor-v3~\cite{nath2020deep} contains $1.5$K images, with just $17$ images including all key classes (``worker,'' ``hard hat,'' and ``vest''), and none showing the ``worker + vest'' combination. Such omissions severely constrain the development of models that necessitate scene and contextual understanding. Furthermore, they lack fine-grained annotations. For example, Pictor-v2~\cite{nath2020deepfrontiers} assigns all machinery a generic ``equipment'' label, limiting the detection of nuanced equipment-specific violations. Such oversimplifications make current models inadequate for practical, real-world deployment. \\
\vspace{-0.75em}

\noindent \textbf{Ignoring Temporal Dynamics.} Current approaches often overlook temporal consistency, which is essential for handling occlusions. Brief occlusions—such as a ``hard hat'' or ``worker'' momentarily hidden—can trigger false alarms. Additionally, when a worker's face is hidden, existing single-view models struggle to detect the person and determine if they are wearing a ``hard hat''.\\
\vspace{-0.75em}

\noindent \textbf{Absence of Standardized Evaluation Metrics.} Despite recent progress, construction violation recognition lacks standardized evaluation metrics. Prior studies~\cite{park2015hardhat, memarzadeh2013automated, luo2018recognizing, kim2018detecting, delhi2020detection, nath2020deep, nath2020deepfrontiers, wu2019automatic, construction-site-safety_dataset} evaluate performance on diverse setups, making direct comparisons difficult. Each defines violation recognition differently, often relying on object detection metrics such as IoU-based mAP~\cite{xiong2021pose, bo2021skeleton, tang2020human}. However, such metrics fall short in assessing a model's ability to correctly identify violations per scene, particularly in complex scenarios involving multiple objects or workers. A more meaningful metric would measure the number of violations correctly identified per scene. Unfortunately, this is not feasible with current models, which are restricted to detecting object classes like ``hard hat'' or ``no hard hat,'' without capturing the violations at the full scene level. To address these gaps, we propose~\modelname, a novel framework for safety violation recognition. Below, we summarize our contributions:

\begin{enumerate}
    \item We are the \textbf{\textit{first}} to formulate violation recognition as a 3D multi-view engagement task. By leveraging geometry-based modeling and multi-view inputs, our approach achieves occlusion-robust, scene-level understanding that surpasses existing state-of-the-art methods.

    \item \modelname{} is the \textbf{\textit{first}} framework to decouple violation criteria from training data, enabling scalable generalization to new violation types without the need for collecting additional real-dataset datasets.
    
    \item We introduce the Synthetic Indoor Construction Scene Generator (SICSG), a novel custom engine that generates physically realistic scene variations, such as illumination, occlusion, and perspective changes, imparting spatial awareness and physical common sense to the model.

    \item We conduct the \textbf{\textit{first}} evaluation in a 3D multi-camera indoor construction setup, comprising four safety violations, four workers and $14$ objects across diverse conditions—occlusions, lighting variations, and camera distances resulting in significant scale changes in worker bodies, significantly increasing scene complexity. \modelname{} consistently outperforms prior methods. Moreover, it is the \textbf{\textit{first}} model tailored specifically for indoor construction settings.
\end{enumerate}
\section{Related Works}
\label{sec:relatedworks}

\noindent\textbf{Sensor-Based Approaches.} Early efforts in construction site safety monitoring relied heavily on sensor technologies, such as RFID tags on safety equipment scanned at site entry points~\cite{naticchia2013monitoring}, LAN-based systems for continuous RFID tracking~\cite{barro2012real}, and short-range transponders paired with wireless networks to verify workers' safety gear~\cite{kelm2013mobile}. While effective in controlled settings, these methods demand substantial infrastructure investment and high maintenance, rendering them impractical for small construction industries. \\
\vspace{-0.75em}

\noindent\textbf{2D Object Detection-Based Approaches.} The complication and high cost of sensor networks spurred interest in vision-based methods, particularly those using affordable RGB cameras. Initial approaches~\cite{park2015hardhat} employed background subtraction to detect workers and assess hard hat usage but struggled with occlusions and static workers, who were often misclassified as background. Advances in RGB-based techniques have since incorporated video cues-\textit{e.g.} motion, color information, edge detection, facial features, and Histogram of Oriented Gradients- for hard hat detection. Notable works by Nath~\textit{et al.}~\cite{nath2020deep, nath2020deepfrontiers} and Delhi~\textit{et al.}~\cite{delhi2020detection} utilized convolutional neural networks (CNNs) to identify basic safety violations. State-of-the-art efforts have predominantly adopted transfer learning with YOLO models, framing safety monitoring as an object detection task~\cite{hayat2022deep,islam2024deep,jia2025geoiou,zhang2025detection,eum2025heavy,ding2025yolo,seth2025enhanced,shetty2024enhancing,chen2021detection,jiao2025detection,zan2025mkd,liu2025improved,onal2024unsafe,park2023small,praveena2025automated,sudharshan2025automated,hwang2025fd,chang2023ffa,jin2024yolo,lian2024hr,yan2022deep,yang2024yolo,cheng2024highly}. Wu~\textit{et al.}~\cite{wu2019automatic} proposed a CNN tailored for hard hat recognition. However, these models often falter in generalization due to scarce construction-specific data~\cite{nath2020deep} and are susceptible to occlusion-induced errors.\\
\vspace{-0.75em}

\noindent\textbf{2D Worker Pose-Based Approaches.} To address object detection limitations, recent studies have explored pose estimation for safety violation recognition. Xiong~\textit{et al.}~\cite{xiong2021pose} pioneered the use of 2D skeleton poses to identify construction safety breaches. Subsequent works, such as Tang~\textit{et al.}, reformulated the problem as a 2D worker-object interaction analysis, while Wang~\textit{et al.}~\cite{bo2021skeleton} applied Graph Convolutional Networks (GCNs) for violation recognition. These methods, however, require retraining for new action classes, limiting scalability. Some approaches have also utilized depth-based cameras (\textit{e.g.} Kinect and VICON) to analyze unsafe worker behaviors. However, their restricted field of view has shifted focus toward RGB camera solutions with broader coverage.\\
\vspace{-0.75em}

\noindent\textbf{Leveraging Vision-Language Models.} Emerging research has leveraged recent vision-language models for safety monitoring. Tsai~\textit{et~al.}~\cite{tsai2025construction} explored fine-tuning contrastive Language-Image Pre-training (CLIP) with prefix captioning to generate automated safety observations. This direction, though promising, is nascent, with state-of-the-art models achieving accuracies of approximately~73.7\%, indicating room for improvement.\\
\vspace{-1em}
\section{Proposed Methodology}

Unlike prior studies, we formulate violation recognition as a 3D multi-view engagement task as shown in Figure~\ref{fig:teaser}. Our methodology estimates 3D worker poses and construction object locations from synchronized multi-view video inputs. Leveraging multiple camera views allows us to effectively address occlusions—an inherent challenge in construction environments—while enabling a comprehensive spatial analysis of worker-object interactions from different camera perspectives.\\
\vspace{-0.75em}

\begin{figure}[t]
    \centering
    \includegraphics[width=\linewidth]{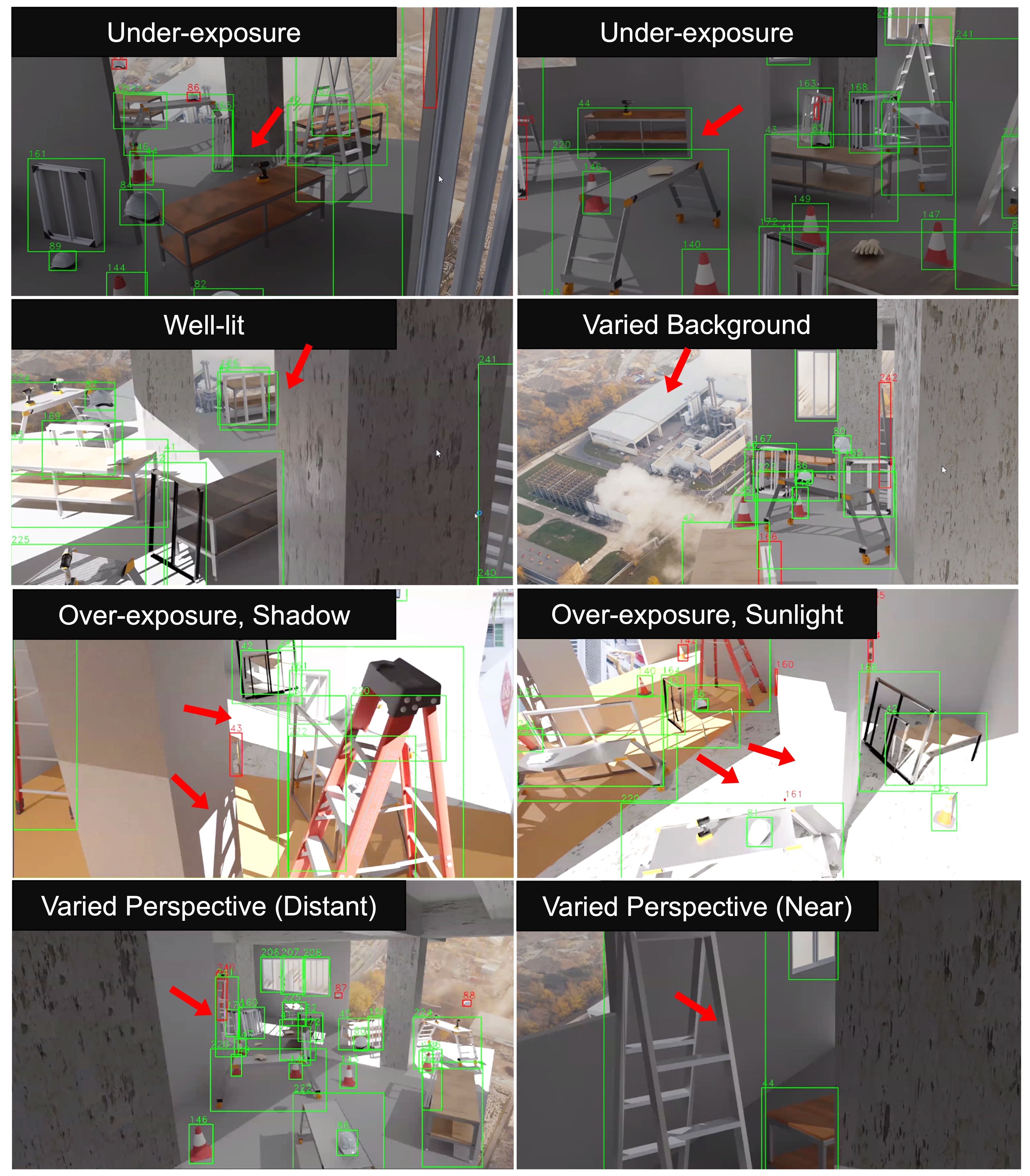}
    \caption{\textbf{SICSG Rendered Data.} Synthetic images generated by SICSG depicting variations in illumination, perspective, and backgrounds, enabling models to exhibit physical commonsense.}
    \label{fig:lighting}
\end{figure}

\noindent\textbf{Synthetic Scene Creation.} To facilitate training of 2D object detectors for construction objects, we introduce the Synthetic Indoor Construction Site Generator (SICSG), developed in Blender 3.1.2. Prior datasets—often crowd-sourced or web-mined—lack sufficient variability and realism for nuanced construction tasks~\cite{nath2020deepfrontiers,nath2020deep,construction-site-safety_dataset}. SICSG overcomes these limitations by generating synthetic datasets tailored specifically to construction tasks, for example, the window installation task for our setting. It features $12$ common object categories across $4000$ images generated from $100$ distinct room arrangements. Systematic variations in lighting, background, and camera perspective (see Fig.~\ref{fig:lighting}) enable models to capture spatial awareness and physical commonsense. For estimating the 2D locations of construction objects, we trained the YOLOv7~\cite{wang2023yolov7} architecture on dataset generated using SICSG.\\
\vspace{-0.75em}

\noindent\textbf{2D Worker and Object Keypoint Estimation.} We employ YOLOv7pose~\cite{wang2023yolov7} to estimate 2D joint positions of workers in each camera view. Worker poses are represented in COCO format as $\mathbf{W}^{i, k}\in\mathbb{R}^{4\times17\times2}$, where $i$ indexes the camera ($1,\dots,N$), and $k$ denotes the detected worker poses per camera view ($1,\dots, K^i$). YOLOv7pose~\cite{wang2023yolov7} is selected for its strong generalization ability in joint estimation tasks. Similarly, construction objects are detected individually per view as $\mathbf{O}^{i, l}$, where $l$ indexes detected objects ($1,\dots,L_i$). To ensure the reliability of keypoint estimations, detections with confidence scores below a threshold $\phi$ are discarded.\\
\vspace{-0.75em}

\begin{figure}[t]
    \centering
    \includegraphics[width=\linewidth]{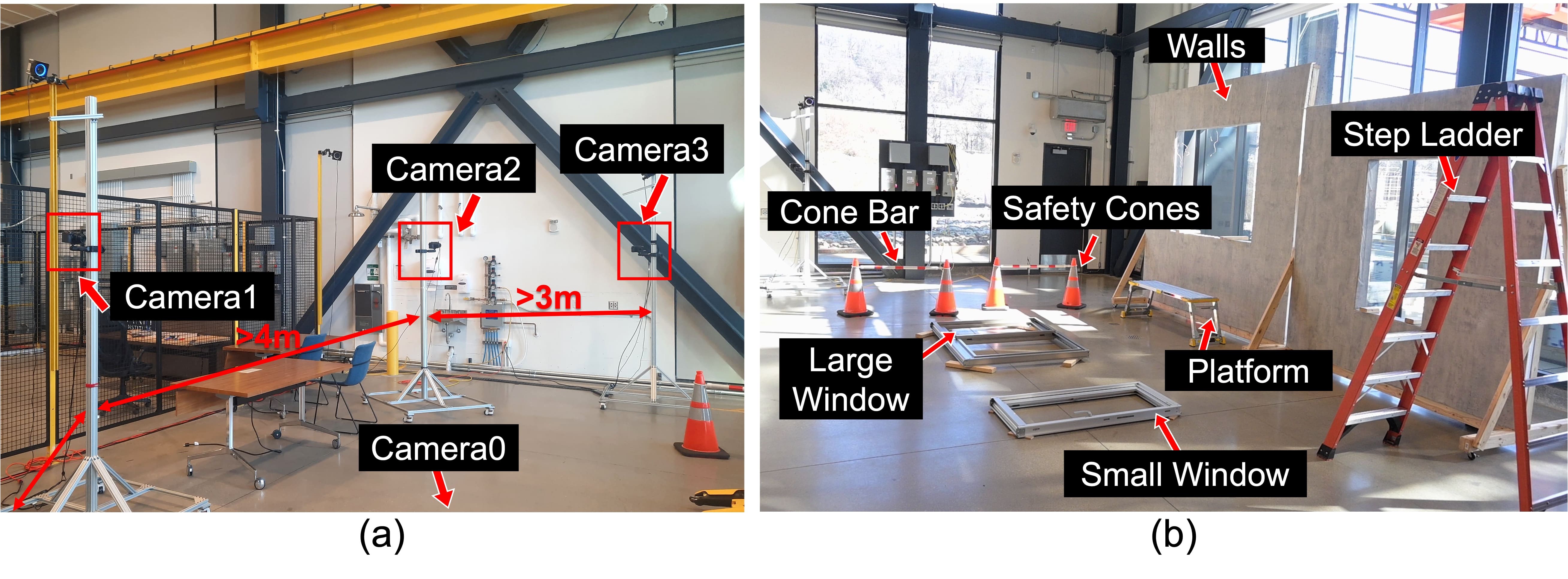}
    \caption{\textbf{Experimental setup.} The evaluation setup consisted of four time-synchronized and calibrated Nikon\textsuperscript{\textregistered} FHD cameras attached to tripods. (a) Cameras 1-3 with Camera 0 out-of-view, (b) Site setting captured by Camera 0.}
    \label{fig:device_settings}
\end{figure}

\noindent\textbf{Cross-View Association via Epipolar Geometry.} Before triangulating into 3D coordinates, the detections from multiple camera views must be accurately associated. For this, we employ epipolar geometry-guided bipartite matching. Initial 2D detections from the first camera view ($i$=$1$) serve as the initial worker and object selections: $\mathbf{H}_s = [\mathbf{H}^{1,k}]$ and $\mathbf{O}_s = [\mathbf{O}^{1,l}]$. 
Subsequently, detections from views $i$=$2,...,N$ are greedily matched with existing worker and object selections using Bi-partite matching. The assignment cost between a new detection ($\mathbf{W}^{i,k}, \mathbf{O}^{i,l}$) and existing selections ($\mathbf{W}^{m}_s, \mathbf{O}^{m}_s$) is computed as follows:
\begin{dmath}
\mathcal{C}_{worker}(\mathbf{W}^{i,k},\mathbf{W}_s^m) = \frac{1}{|\mathbf{W}_c^m||J_{kl}|} \sum_{\mathbf{W}^{j,l}\in\mathbf{W}_s^m} \sum_{\mathbf{W}^{j,l}\in\mathbf{J}_{kl}} \mathcal{D}(\mathbf{W}^{i,k}(\varphi),\mathbf{W}^{j,l}(\varphi))
\end{dmath}
\vspace{-8mm}
\begin{dmath}
    \mathcal{C}_{object}(\mathbf{O}^{i,k},\mathbf{O}_s^m) = \frac{1}{|\mathbf{O}_s^m||J_{kl}|} \sum_{\mathbf{O}^{j,l}\in\mathbf{O}_s^m} \sum_{\mathbf{O}^{j,l}\in\mathbf{J}_{kl}} \mathcal{D}(\mathbf{O}^{i,k}(\varphi),\mathbf{O}^{j,l}(\varphi))
\end{dmath}
\vspace{-3mm}
where $\mathbf{W}^{i,k}(\varphi)$ denotes the 2D-pixel location of joint $\varphi$ of the 2D worker $\mathbf{W}^{i,k}$ and $J_{kl}$ is the set of joints that are visible and non-occluded for both poses $\mathbf{W}^{i,k}$ and $\mathbf{W}^{j,l}$. The same is extended to objects with the number of joints always being one. The distance between two joints in the respective cameras is defined by the distance between the epi-polar lines and the joint locations:

\begin{equation}
    \mathcal{D}(p_i, p_j) = |p_j^T F^{i,j} p_i| + |p_i^T F^{j,i} p_j|
\end{equation}
where $F^{i,j}$ is the fundamental matrix from Camera $i$ to Camera $j$. Solving the bipartite matching problem
\begin{equation}
    X^* = \operatorname*{min}_X \sum_{m=1}^{|\mathbf{W}_s^m|} \sum_{k=1}^{K_i}C(\mathbf{W}^{i,k}, \mathbf{W}_s^m)X_{k,m}
\end{equation}
\begin{equation}
    \sum_k X_{k,m} = 1 \ \forall \ m \ \text{and} \ \sum_m X_{k,m} = 1 \ \forall \ k
\end{equation}
$X^*_{k,m} = 1$ if $\mathbf{W}^{i,k}$ is associated with an existing worker selection, else $0$. If $X^*_{k,m} = 1$ and $\mathcal{C}_{worker}(\mathbf{W}^{i,k},\mathbf{W}_s^m) < \theta$, a match is confirmed and the 2D detection $\mathbf{W}^{i,k}$ is added to $\mathbf{W}_s^m$. Otherwise, the detection is treated as a new entity and $\mathbf{W}^{i,k}$ is added as selection for a new worker to $\mathbf{W}$. This process is repeated for objects.\\
\vspace{-0.75em}

\noindent\textbf{3D Triangulation.}
Finally, we triangulate the associated detections using calibrated camera parameters. For construction objects, we compute a weak pose by triangulating the bounding box center, suitable for small items (\textit{e.g.}, hard hat) occupying $<$$1\%$ of image pixels (see Fig.~\ref{fig:results}). Despite minimal pixel coverage, this method proves reliable for 3D localization, as small objects can be approximated as points in 3D space.\\
\vspace{-0.75em}

\noindent\textbf{Bipartite Matching for Tracking.} To track the entities over time, we employ bipartite matching~\cite{munkres1957algorithms}. The newly triangulated 3D worker and weak-object pose at time $t$ are matched to the previous 3D poses at time $t$-$1$ by bipartite matching. The matching cost is computed as the Euclidean distance between all corresponding joints that are present in both poses. If this cost is below a predefined threshold (denoted as $\psi$), the poses are considered matched; otherwise, they are treated as new workers or objects. In cases where two consecutive poses lack joints or have missing joints, the mean of all present joints is calculated. When computing the cost under such circumstances, the points are projected onto the $x$-$y$ plane before measuring the Euclidean distance between them.\\
\vspace{-0.75em}

\begin{figure}[t]
    \centering
    \includegraphics[width=\linewidth]{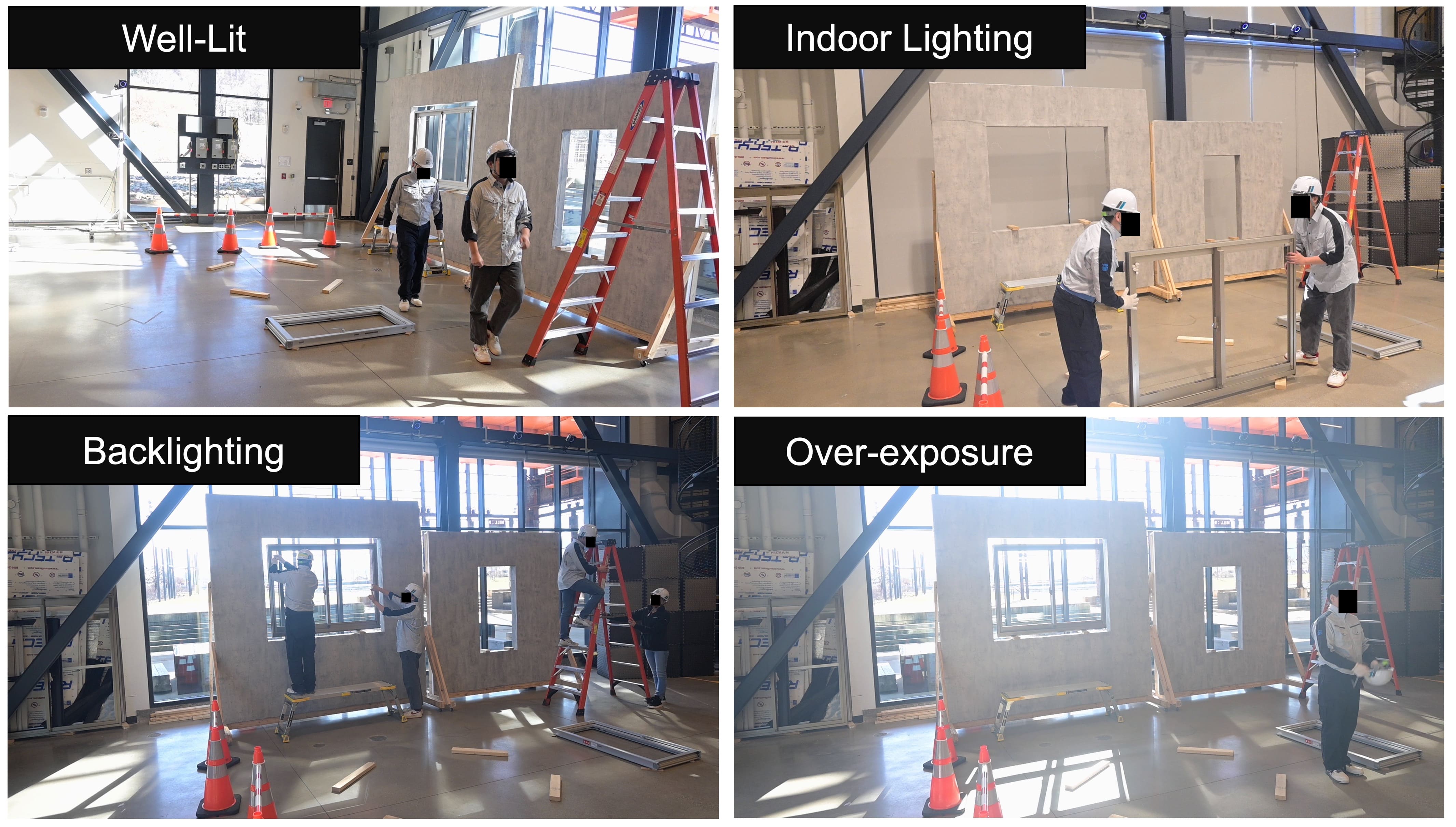}
    \caption{\textbf{Illumination Variations.} Sample lighting scenarios encountered during testing of \modelname{}. The first row shows well-lit indoor rooms common in indoor construction; the second row presents backlighting and over-exposure, which reduces visibility and presents a darker appearance.}
    \label{fig:lighting_supplementary}
\end{figure}

\begin{figure}[t]
    \centering
    \includegraphics[width=\linewidth]{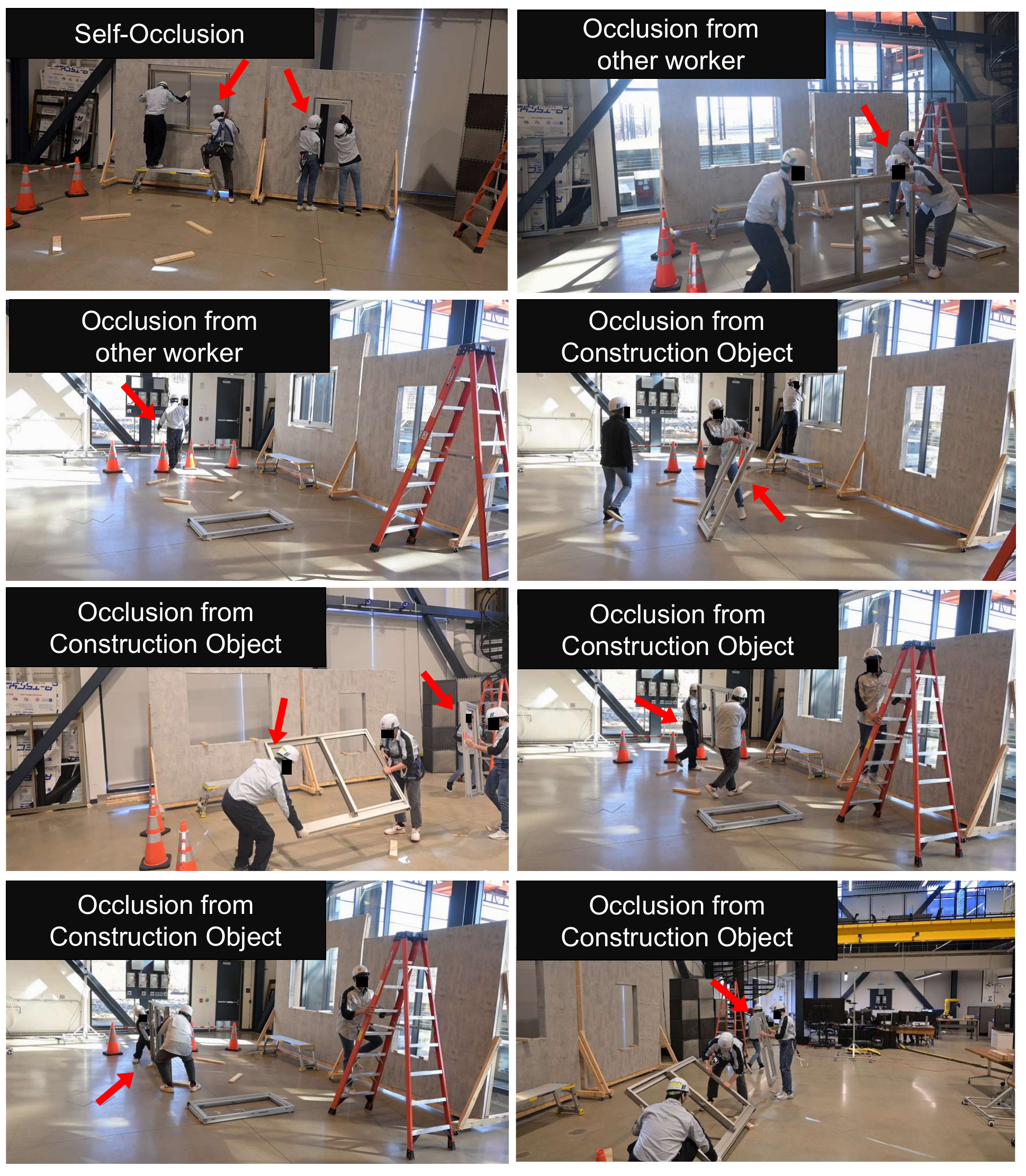}
    \caption{\textbf{Occlusions Variations.} Examples of self-occlusion and secondary occlusion during \modelname{} testing.}\label{fig:occlusion_supplementary}
\end{figure}

\begin{table*}
    \setstretch{1}
    \begin{threeparttable}
    \begin{adjustbox}{max width=\linewidth}
    \begin{tabular}{llcccclc}
    \toprule
    \multirow{2}{*}{Methodology} & \multirow{2}{*}{Dataset} & \multicolumn{4}{c}{Testing Set Specifications} & \multicolumn{2}{c}{Violation Recognition}\\
    \cmidrule(lr){3-6}
    \cmidrule(lr){7-8}
    & & \multirow{1}{*}{Collection Method} & \multirow{1}{*}{Type} & \multirow{1}{*}{Scene-Level} & \multirow{1}{*}{Evaluation Metric} & \multirow{1}{*}{Classes} & \multirow{1}{*}{\#Violations$^{\textcolor{blue}{\dag}}$} \\
    \midrule
    \multicolumn{8}{l}{\textbf{Object Detection-based Formulation}}\\
    \midrule
    
    BGR Subtraction & Park \textit{et al.} \cite{park2015hardhat} & Site & Video & - & Confusion Matrix & Hardhat, Worker & 1\\
    
    CNN & GDUT-HWD \cite{wu2019automatic} & Web & Image & {\color{red}\xmark} & IoU mAP & Hardhat, No hardhat & 1\\ 
    
    YOLOv3 Transfer Learning & Delhi \textit{et al.} \cite{delhi2020detection} & Site, Web & Image & - & Confusion Matrix & 4 Safety Items & 2\\
    
    YOLOv2-v3 Transfer Learning & Pictor-v2 \cite{nath2020deepfrontiers} & Crowd, Web & Image & {\color{red}\xmark} & IoU mAP & 3 Site Elements & -\\
    
    YOLOv3 Transfer Learning & Pictor-v3 \cite{nath2020deep} & Crowd, Web & Image & {\color{red}\xmark} & IoU mAP, Confusion Matrix & 3 PPE Types & 2\\
    
    YOLOv8 Transfer Learning & Roboflow~\cite{construction-site-safety_dataset} & Crowd, Web & Image & {\color{red}\xmark} & IoU mAP, Confusion Matrix & 10 Site Elements & 2\\

    \midrule
    \multicolumn{8}{l}{\textbf{2D Human Pose/ Object Interaction}}\\
    \midrule

    Detection-based HOI & Tang \textit{et al.} \cite{tang2020human} & Crowd, Web & Image & - & IoU mAP & 22 Categories & 4\\
        
    Part-aware Localization & CPPE~\cite{xiong2021pose} & Web & Image & {\color{red}\xmark} & IoU mAP & 7 PPE States & 3\\
    
    GCN & Wang \textit{et al.}~\cite{bo2021skeleton} & Site & Video & - & Confusion Matrix & 6 Behaviors & 5\\

    \midrule
    \multicolumn{8}{l}{\textbf{3D Multi-view Engagement Task Formulation}}\\
    \midrule
    
    \modelname{} (Ours) & SICSG & Synced Setup & Video & {\color{green}\cmark} & Scene-level Confusion Matrix & 4 Dynamic Cases & 4\textcolor{blue}{\ddag}\\
    \bottomrule
    \end{tabular}
    \end{adjustbox}
    \vspace{-2mm}
    \caption{\textbf{Comparison with Prior Methods.} \modelname{} compared to existing models (we do not consider RGBD). $^{\textcolor{blue}{\dag}}$$\pm$ Categories of same violation counted once $^{\textcolor{blue}{\ddag}}$ Evaluated on four cases. \modelname{} violations are not dataset-dependent.}
    \label{tab:OverviewDatasets}
    \end{threeparttable}
\end{table*}

\noindent\textbf{Compliance and Violation Queries.} The resulting 3D worker and object pose facilitate straightforward safety compliance verification through simple geometric queries based on L2-norm. For \textit{e.g.}, four typical safety violations are identified using thresholds derived from worker body dimensions as shown below:\\
\vspace{-2mm}
\begin{itemize}
    \item No hard hat:
    \vspace{-2mm}
    \begin{equation}
    ||\mathbf{W}_{neck} -\mathbf{O}_{hardhat}||_{L2} < \tau_1
    \end{equation}

    \item Single-worker platform restriction:
    \vspace{-2mm}
    \begin{equation}
        ||\mathbf{W}_{feet} -\mathbf{O}_{platform}||_{L2} < \tau_2
    \end{equation}

    \item Ladder holding requirement:
    {\footnotesize \begin{equation}
        ||\mathbf{W}^{(1)}_{torso} -\mathbf{O}_{ladder}||_{L2} < \tau_3, ||\mathbf{W}^{(2)}_{torso} -\mathbf{O}_{ladder}||_{L2} < \tau_3
    \end{equation}}
    \vspace{1mm}

    \item Large window requires dual-worker handling:
    {\footnotesize \begin{equation}
        ||\mathbf{W}^{(1)}_{torso} -\mathbf{O}_{window}||_{L2} < \tau_4, ||\mathbf{W}^{(2)}_{torso} -\mathbf{O}_{window}||_{L2} < \tau_4
    \end{equation}}
    \vspace{1mm}
\end{itemize}

\noindent Thresholds $\tau_1,\dots,\tau_4$ are empirically derived and scaled relative to standard worker body dimensions. For example, $\tau_1$ is set as one-tenth of the worker height $W_{height}$. Since the camera coordinate system normalizes scale, these thresholds are effective and invariant across different worker body sizes.
\vspace{-3mm}
\section{Experiments and Results}

\subsection{Experimental Setup}

In this section, we present the experimental setup, results, and ablation studies for the safety violation recognition task in construction environments.\\
\vspace{-0.75em}

\noindent \textbf{Camera Setup and Synchronization.} The experimental configuration involved up to four workers, $14$ objects, and four safety violations, captured by four synchronized and calibrated Nikon FHD RGB cameras (indexed $0$-$3$), each with a resolution of $1920$$\times$$1080$ at $30$ FPS. To ensure stability, all cameras were mounted on tripods. The target area was designed to emulate a real-world construction site by maximizing its size while maintaining a converging layout. As illustrated in Fig.~\ref{fig:device_settings}, the cameras were positioned in a semi-circle around the target area at a height of $2.0$ meters. Distances from the cameras to the scene center ranged from $3$-$10$ meters, enabling the capture of large-scale scenes with significant variations in worker scale. Synchronization was achieved via a wired interface and connected to a central server, with time offsets computed between each camera and the server's clock to ensure precise temporal alignment, following prior motion capture studies~\cite{cai2022humman, h36m_pami, Joo_2017_CMU_TPAMI}. Timestamp-based file naming was followed to store temporal data.

\begin{figure*}[t]
    \centering
    \includegraphics[width=\linewidth]{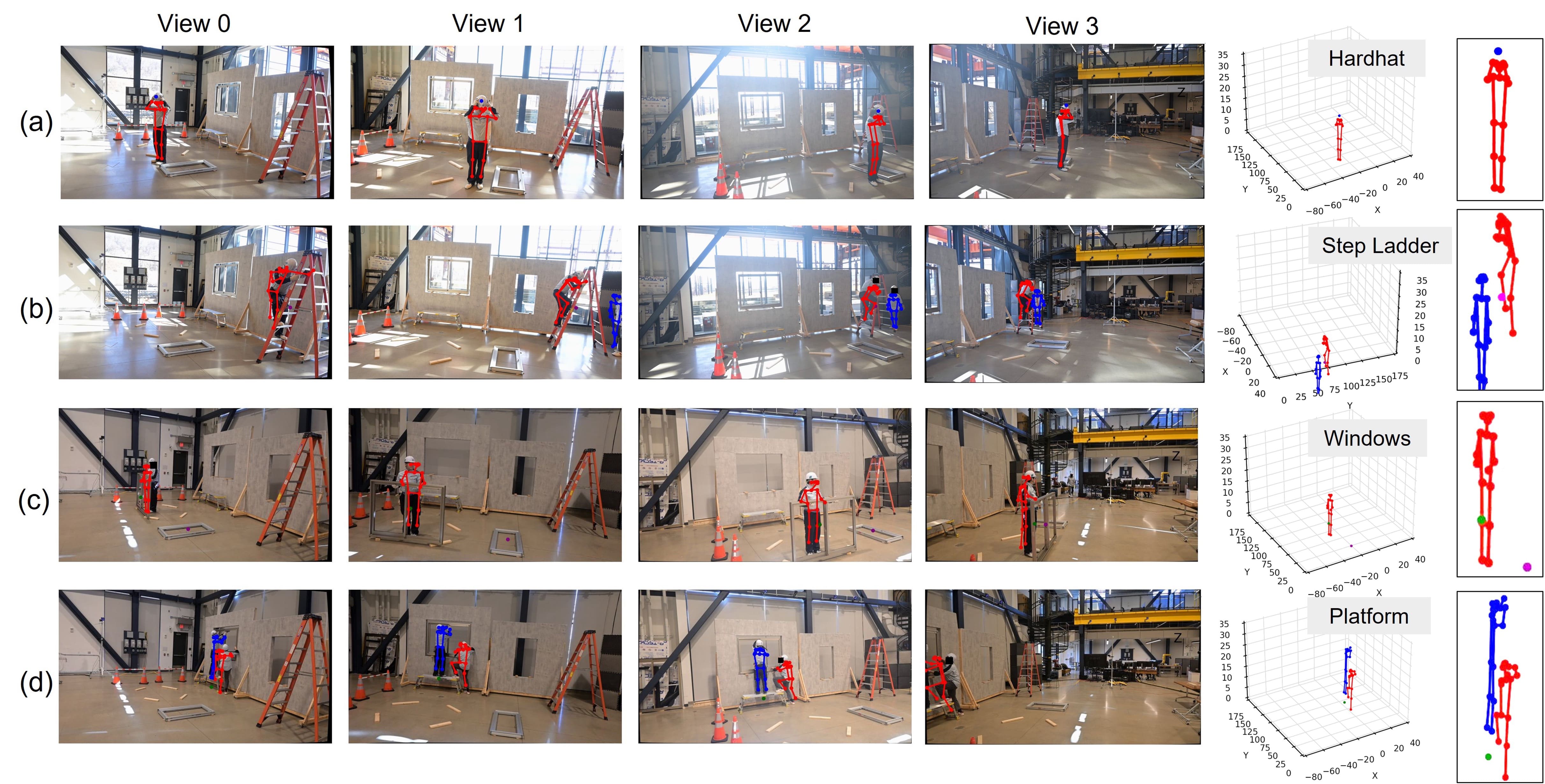}
    \captionsetup{font={small}}
    \vspace{-1.5em}
    \caption{
    \textbf{Safety Violation Recognition Qualitative Results.} We show the 2D re-projection of worker and object's pose on the image plane: Rows (a), (b) show two safe scenarios, while (c), (d) illustrate two violations: (a) Worker wearing a {\color{blue} hard hat}, (b) Second worker holding the {\color{magenta} Step ladder} when the first climbs it, (c) Only one worker is carrying a {\color{teal} Large window} that should be carried by two workers showing a violation scenario (the {\color{magenta} small window} is represented in magenta), (d) Two workers are standing on the {\color{teal} Platform} simultaneously.
    }
    \label{fig:results}
    \vspace{1em}
\end{figure*}

\begin{table*}
    \centering
    \setstretch{1.1}
    \begin{adjustbox}{max width=0.95\linewidth}
        \begin{tabular}{llc|c|c|cc}
        \toprule
        \multirow{3}{*}{Models} &\multirow{3}{*}{Supervision} & \multicolumn{4}{c}{Safety Violation Recognition Accuracy (\%)} & \multirow{3}{*}{AVG.}\\
        \cline{3-6}
        &  & \multirow{2}{*}{No Hard hat} & \multirow{1}{*}{Ladder not stabilized by the second} & \multirow{1}{*}{Large Windows requires} & \multirow{1}{*}{Only one worker at a time is} & \\
        &  &  & \multirow{1}{*}{worker when the first climbs} & \multirow{1}{*}{handling by two workers} & \multirow{1}{*}{allowed on the platform} & \\
        \midrule
        \multirow{1}{*}{Baseline} & \multirow{1}{*}{One view} & 81.7 & 78.2 & 84.5 & 91.4 & -\\
        \hline
        \multirow{4}{*}{\textbf{Ours}} & Two views & 84.0 & 81.6 & 87.2 & 91.8 & \multirow{2}{*}{\textbf{+2.2\%}}\\ 
        & Improvement (abs) & \cellcolor{green!50}$\textbf{+2.3\%}$ & \cellcolor{green!30}$\textbf{+3.4\%}$ & \cellcolor{green!20}$\textbf{+2.7\%}$ & \cellcolor{green!20}$\textbf{+0.4\%}$ & \\
        \cline{2-7}
        & Four views & 92.0 & 88.4 & 91.4 & 94.2 & \multirow{2}{*}{\textbf{+7.6\%}}\\
        & Improvement (abs) & \cellcolor{green!50}$\textbf{+10.3\%}$ & \cellcolor{green!50}$\textbf{+10.2\%}$ &\cellcolor{green!30}$\textbf{+6.9\%}$& \cellcolor{green!20}$\textbf{+2.8\%}$ & \\
        \bottomrule
        \end{tabular}
    \end{adjustbox}
    \caption{\textbf{Quantitative Comparison.} Performance for the violation recognition task. Values are rounded to the first decimal place.} 
    \label{tab:results_final}
\end{table*}

\noindent Camera parameters were estimated using checkerboards following OpenCV standards~\cite{opencv_library, CalibrationZhang}, with Camera $0$ as the reference origin. Due to the large target area, checkerboards were not simultaneously visible across all four cameras, necessitating a pairwise calibration approach for extrinsics. Intrinsic calibration yielded low RMSE values of $0.2567$, $0.1823$, $0.2092$, and $0.1623$ (pixels) for Cameras $0$-$3$, and an extrinsic RMSE of $0.8098$, $0.5806$, and $1.1369$ for Cameras $1$-$3$ relative to Camera $0$.\\
\vspace{-0.75em}

\noindent\textbf{Illumination and Occlusion Variations.} Real-world construction sites are characterized by diverse lighting conditions and frequent occlusions. Figures~\ref{fig:lighting_supplementary} and~\ref{fig:occlusion_supplementary} illustrate the test setup, which represents these challenges. Lighting variations included indoor illumination, back-lighting scenarios with sunlight exposure, and over-exposure conditions. Occlusions ranged from partial or complete obstruction of workers by construction objects to self-occlusions caused by intricate worker movements or poses during tasks.\\
\vspace{-0.75em}

\noindent\textbf{Safety Violation Scenarios.} We evaluated four intricate and complex real-world safety violations specific to indoor window installation. This included: (1) No hard hat, (2) Step Ladder not stabilized/ held by a second worker when the first worker climbs it, (3) Carrying a large window requires dual handling, (4) Only one worker at a time is allowed on a platform. Unlike previous models, scene-level annotations were applied consistently across all frames of a sequence, even if a violation was occluded in some views. For instance, if a safety violation occurred, all four frames were labeled with that violation as ground truth, even if the violation was not visible due to occlusion in one of the frames. The model was implemented on an NVIDIA RTX-3090 GPU with 64GB RAM.\\
\vspace{-0.75em}

\begin{figure*}
    \centering
    \includegraphics[width=\linewidth]{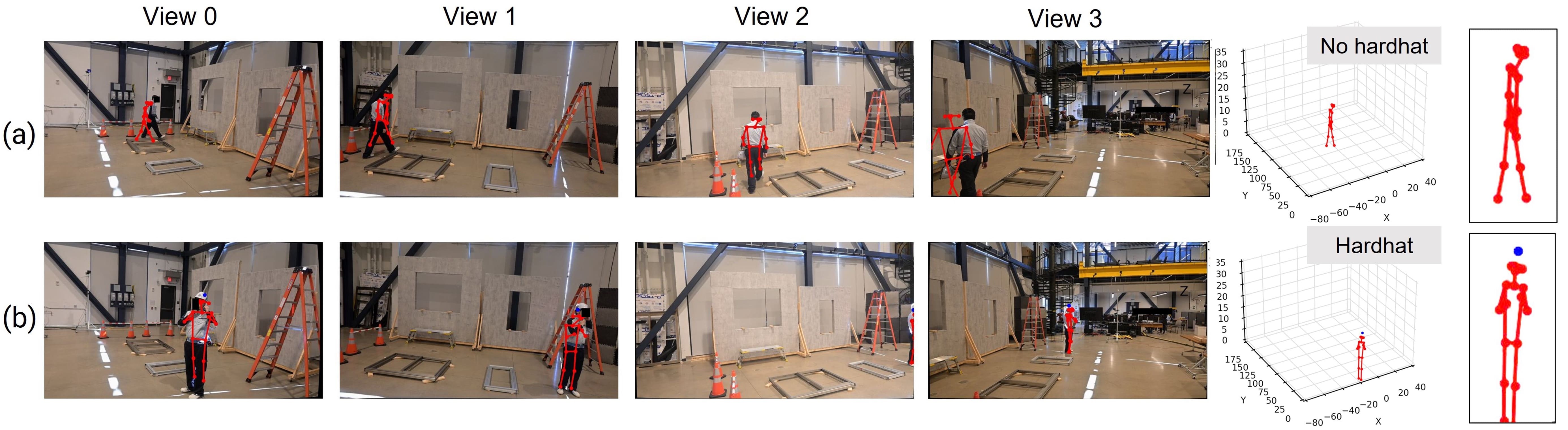}
    \captionsetup{font={small}}
    \vspace{-1em}
    \caption{\textbf{Edge Cases.} (a) A case when the hard hat is not detected. We mark this as a violation, based on the previous frame \textit{i.e.}, unless the worker wears the hard hat again, all frames are tagged as violations. (b) Increasing the number of views improves model prediction.
    }
    \label{fig:results_failures}
\end{figure*}

\begin{figure}
    \centering
    \includegraphics[width=\linewidth]{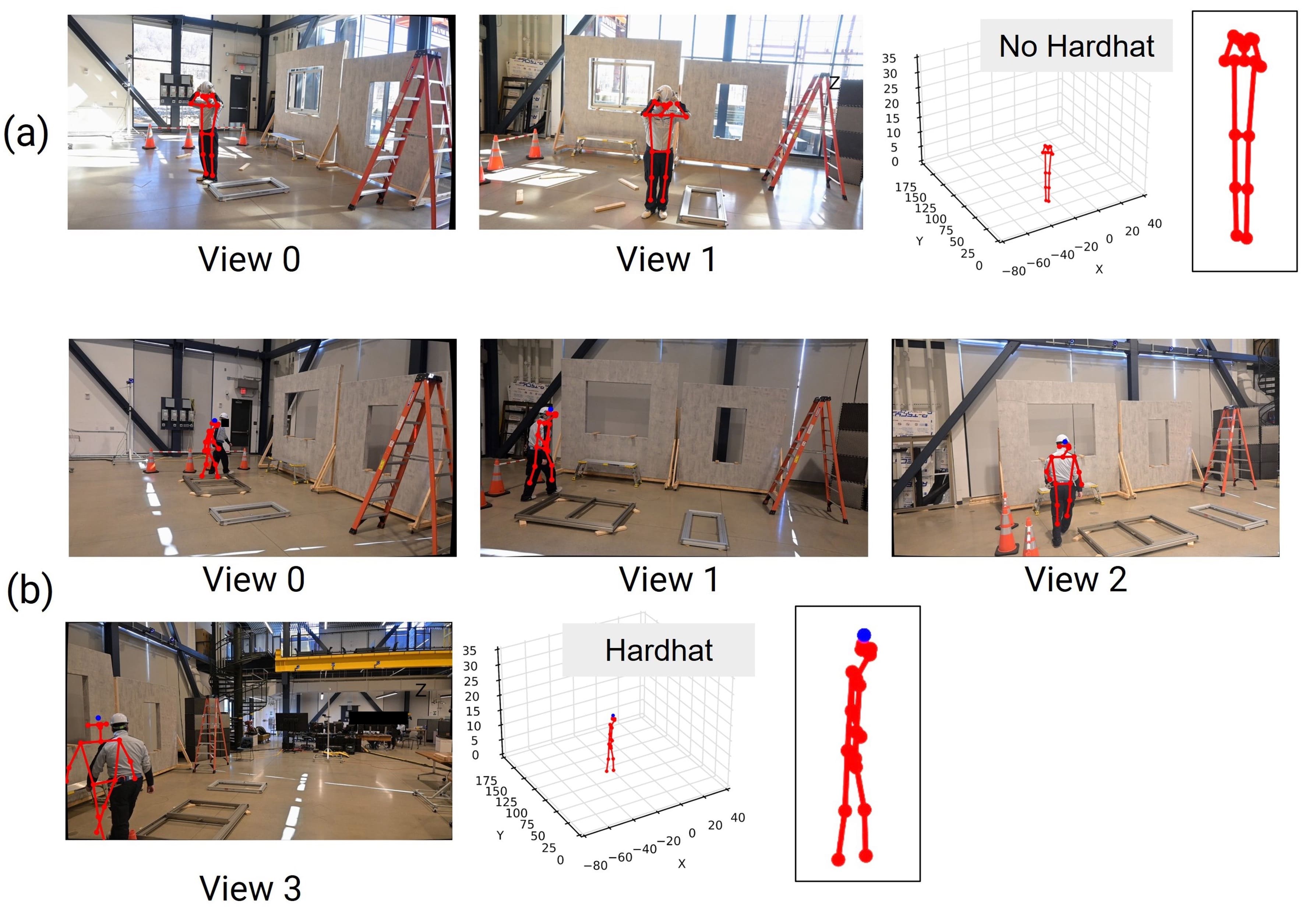}
    \captionsetup{font={small}}
    \caption{\textbf{Failure Cases.} (a) Results using $2$-views highlighting missed hard hat prediction. (b) Pose distortions.}
    \label{fig:results_ablations}
\end{figure}

\noindent \textbf{Why Accurate Pose Estimation is Not the Focus?} We do not compare our model's results with multiview 3D human pose estimation benchmarks, such as CMU-Panoptic~\cite{Joo_2017_CMU_TPAMI} or Human3.6M~\cite{ionescu2013human3}, as these are primarily intended for 3D human pose studies without containing construction objects or safety violation recognition. Moreover, pose distortions do not critically impair violation recognition (see Figure~\ref{fig:results_ablations}) as geometric triangulation suffices for our use case. In our use case, real-time pose estimation is prioritized over achieving an extremely accurate pose because timely recognition is critical for practical applications in construction environments. Therefore, geometric triangulation was employed without additional methods to enhance pose accuracy, which is out of the scope of the current study.

\subsection{Results and Discussion} 

We test our model for four violations in the challenging over-exposure case. For comparison with previous approaches, we re-trained the current SOTA model on the same synthetic dataset from SICSG. Herein, we designed a 2D single-frame violation recognition model using the object-detection-based approach. For a fair comparison, we used the same backbone model and L2-norm-based safety criteria and thresholds. Further, to make the comparison even more challenging, we utilized camera $i$=$1$, which offered the front-view and is the most advantageous. This is also consistent with industrial setups, wherein the front camera is the main input. Table~\ref{tab:results_final} reports recognition accuracy across the four violation types. The baseline achieves 81.7–91.4\% accuracy, while our method with two views improves performance by an average of +2.2\%, and with four views by +7.6\%. Notable gains include +10.3\% for ``No hard hat'' and +10.2\% for ``Ladder Not Held'', highlighting the advantage of multi-view supervision. In summary, our multi-view system significantly outperforms single-view baselines, with a four-view setup achieving up to +10.3\% accuracy gains. These results validate the efficacy of multi-camera supervision for safety violation recognition in complex, occluded environments. We present the qualitative results in Figures~\ref{fig:results}, ~\ref{fig:results_failures}, and ~\ref{fig:results_ablations}, including limitations and failure cases.\\
\vspace{-1em}

Figure~\ref{fig:results} illustrates 2D re-projections of worker and object poses, with rows (a)–(b) depicting safe scenarios (\textit{i.e.}, worker with hard hat, ladder held) and rows (c)–(d) showing violations (\textit{i.e.}, single worker with large window, crowded platform). Edge cases (see Figure~\ref{fig:results_failures}), such as missed hard hat recognition due to occlusion, are mitigated by temporal consistency and multi-view data.\\
\vspace{-0.75em}

\noindent\textbf{Effect of Multiple-view 3D Spatial Understanding.} To assess the impact of multiple views, compared to single-camera setups, model performance was evaluated across varying supervision (Table~\ref{tab:results_final}). Two views enhance violation recognition over the baseline (\textit{e.g.}, +3.4\% for ``Ladder Not Held''), while four views yield the highest gains, particularly in occlusion-heavy scenarios. Results indicate that increasing the number of views from one to four improves average performance by $+7.6\%$. This underscores the robustness of a multi-view setup for addressing occlusions.
\vspace{-2mm}
\section{Conclusion and Future Work}

We present~\modelname, a novel 3D violation recognition model as a step towards addressing the limitations of current models in multi-view indoor construction environments. Experimental results demonstrate~\modelname's superiority to previous methods. We believe that \modelname{} will catalyze the development of algorithms designed to sense violations on construction sites, leading to improvement in safety in these industrial environments.

One constraint of \modelname{} lies in its reliance on precise camera calibration. To emulate the real world, our setup consisted of a large target area. We utilized pairwise stereo calibration to calibrate the cameras. However, this has a compounding effect on RMSE values across multiple pairwise calculations. If not calibrated precisely, this can adversely affect the model's performance. Additionally, motion blur—stemming from the 30 FPS capture rate—poses a challenge, as high-frame-rate cameras are uncommon in real-world construction settings. Thus, developing algorithms resilient to motion blur is critical. Other avenues for future work include: \textit{(i)} analyzing the trade-off between violation recognition accuracy and precision of 3D joint estimation for workers, and \textit{(ii)} optimizing runtime performance by retraining a pose estimation backbone that focuses solely on the worker joints that are relevant to the violation criteria.\\
\vspace{-0.5em}

\noindent \textbf{Acknowledgments.} The work was supported by YKK-AP Inc., Japan. Aviral Chharia was supported in part by the ATK-Nick G. Vlahakis Graduate Fellowship from Carnegie Mellon University, Pittsburgh. The authors are thankful to Aman Chulawala, Hidetaka Kajikawa, Naoto Tanaka, and Yoriko Ogawa for helpful discussions and feedback to improve the work. Work was done while Tianyu Ren was at Carnegie Mellon University.\\
\vspace{-1em}
{
    \small
    \bibliographystyle{ieeenat_fullname}
    \bibliography{main}
}
\end{document}